\documentclass[11pt]{article}
\usepackage[preprint]{acl}
\usepackage{times}
\usepackage{latexsym}
\usepackage[T1]{fontenc}
\usepackage[utf8]{inputenc}
\usepackage{microtype}
\usepackage{inconsolata}
\usepackage{graphicx}
\usepackage{booktabs}
\usepackage{amsmath}
\usepackage{amssymb}
\usepackage{multirow}
\usepackage{array}
\usepackage[capitalize,noabbrev]{cleveref}
\usepackage{listings}
\usepackage{authblk}
\lstdefinestyle{promptstyle}{
  basicstyle=\ttfamily\scriptsize,
  breaklines=true,
  frame=l,
  numbers=left,
  numberstyle=\tiny,
  numbersep=6pt,
  aboveskip=6pt,
  belowskip=6pt,
  xleftmargin=2em,
  framexleftmargin=1.5em,
  columns=flexible,
}

\title{CobSeg: Coherence Boundary Modeling for Dialogue Topic Segmentation}

\author{
\textbf{Sijin Sun}$^{1}$ \quad
\textbf{Liangbin Zhao}$^{1,*}$ \quad
\textbf{Jiaxiang Cai}$^{1}$ \quad
\textbf{Ming Deng}$^{2}$ \quad
\textbf{Mingyu Luo}$^{3}$ \quad
\textbf{Xiuju Fu}$^{1}$
}
\affil{\normalsize
$^*$Corresponding author \\
$^1$ Institute of High Performance Computing, Agency for Science, Technology and Technology \\
$^2$ Shanghai Univeristy \quad
$^3$ Fudan University
}

\begin{document}
\maketitle
\begin{abstract}
Dialogue topic segmentation is critical in many human-AI collaborative applications which requires identifying heterogeneous boundary cues, including lexical transitions near utterance edges and semantic discontinuities across utterances. Existing utterance models often dilute these local lexical signals. We propose CobSeg, a novel multi-branch architecture that separates coherence-level semantic continuity from lexical boundary transitions and recovers both through directional boundary prediction. CobSeg further uses boundary informativeness weighting to emphasize high-utility utterance positions, and incorporates a corpus-derived topic coherence cue with learned combination weights. While CobSeg is evaluated as a compact trainable segmenter under supervised gold-boundary training and a pseudo-label setting with automatically induced boundaries, it performs enhanced boundary prediction without LLM calls during inference. Across five benchmarks, it improves $P_k$ and $W_d$ particularly when local lexical cues are prominent: under gold supervision, it reduces $P_k$ by 0.7 points and $W_d$ by 0.6 points on VHF, and reaches $P_k$ of 1.0 on DialSeg711; with induced boundaries, it reduces $P_k$ by 14.8 points on VHF, by 1.5 points on DialSeg711, and by 1.1 points on TIAGE, outperforming prior non-LLM approaches.
\end{abstract}

\section{Introduction}

\begin{figure}
    \centering
    \includegraphics[width=1.0\linewidth]{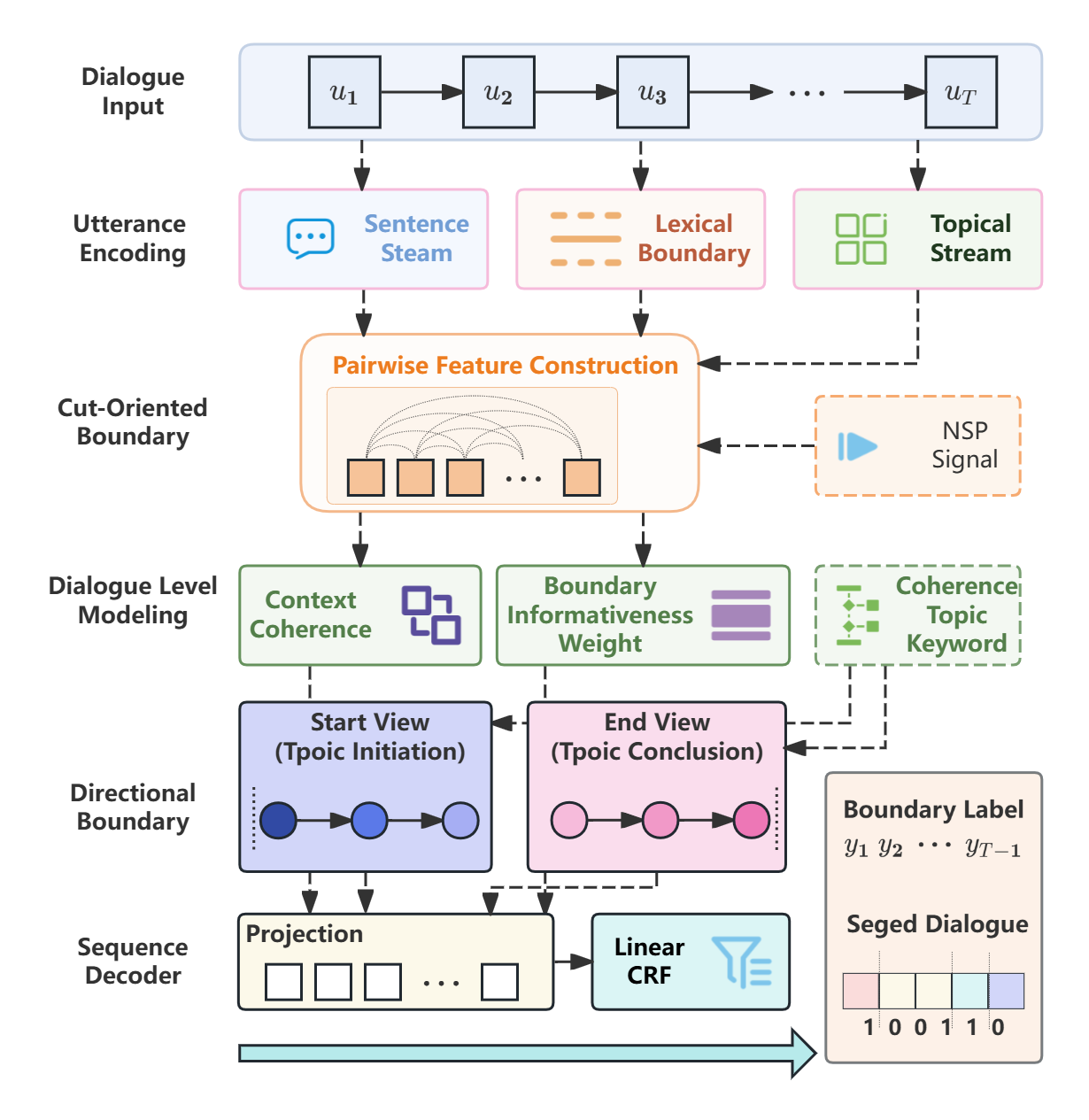}
    \caption{Overview Structure of CobSeg.}
    \label{fig:summary_structure}
\end{figure}
Dialogue topic segmentation identifies topic shifts in multi-turn conversations and partitions dialogues into coherent topical units, making long and unstructured conversations easier to organize and process in practical applications such as customer service analysis, meeting review, and traffic communication, monitoring and management. It supports downstream tasks including dialogue understanding \citep{wang2020response}, retrieval \citep{xu2021discovering}, and summarization \citep{qi2021improving}. Especially, it helps enhance human–AI collaboration in Vessel Traffic Services (VTS) and Air Traffic Control (ATC) systems. Recent work spans supervised models trained on gold boundaries \citep{koshorek2018text,jiang2023superdialseg}, unsupervised methods that induce boundaries from coherence patterns \citep{gao2023unsupervised,xing2021improving}, pseudo-label training with auxiliary signals \citep{artemiev2024leveraging}, and LLM-based reasoning \citep{lee2025defdts,das2024s3}. 

However, direct LLM inference represents a different operating regime from trainable segmentation models. LLM-based methods typically rely on large general-purpose models and prompt-based reasoning at test time, while trainable segmenters aim to learn task-specific boundary predictors that can be deployed independently after training. This distinction is important in practical settings where inference cost, latency, reproducibility, on-premise deployment requirements, and data privacy make repeated LLM calls less desirable.

As \Cref{fig:summary_structure} illustrates, the core challenge is assigning reliable decision signals to candidate boundaries by integrating utterance-level content, cut-level transition evidence, and sequence-level decoding.

Existing work models DTS tasks from the utterance level, which introduces limitations. \citep{devlin2019bert,liu2019roberta} compress sentence-level encoders under each utterance into a fixed vector, averaging boundary-adjacent tokens with mid-utterance content and diluting the lexical transitions that mark topic shifts. Recent utterance-pair frameworks \citep{yang2025unified,somasundaran2020two} operate entirely at the utterance level and cannot recover these token-level cues. \citep{li2018segbot,nair2023neural} allocate capacity uniformly across all positions, while only a small fraction of utterance positions carry decisive transition signals. \citep{gao2023unsupervised,park2023unsupervised,glavavs2016unsupervised,gong2022tipster} use corpus-level topic coherence to fixed preprocessing heuristics rather than understanding the deep pattern of multi-utterance pairing.

Proposed CobSeg addresses each limitation through a targeted design, while following the trainable-segmenter regime described above: it performs boundary prediction with a compact task-specific model and does not require LLM calls during inference. A Lexical Boundary Detector retains token-level transition evidence by upweighting utterance-edge tokens before pooling. Utterance boundary informativeness weighting (UBIW) learns per-position informativeness scores, concentrating capacity on high-utility cut positions. A topic coherence cue from unsupervised keyword induction supplies statistical boundary evidence with learned coefficients, injected at the logit level so the model can strengthen or suppress the signal during training. Directional boundary heads separately model topic conclusion and topic initiation, reflecting the asymmetry of the two signals.

The main contributions are:
\begin{itemize}
\item We propose CobSeg, a compact trainable multi-branch framework that separates lexical transition cues from semantic coherence signals for DTS, enabling efficient boundary prediction without inference-time LLM calls.

\item We introduce UBIW and directional prediction heads to model sparse, asymmetric evidence around topic shifts.

\item We evaluate CobSeg under supervised and pseudo-label settings on five benchmarks, showing improvements in $P_k$ and $W_d$ with targeted ablations.
\end{itemize}

\section{Methodology}

\subsection{Overall Structure}

\begin{figure*}[!t]
    \centering
    \includegraphics[width=1.0\textwidth]{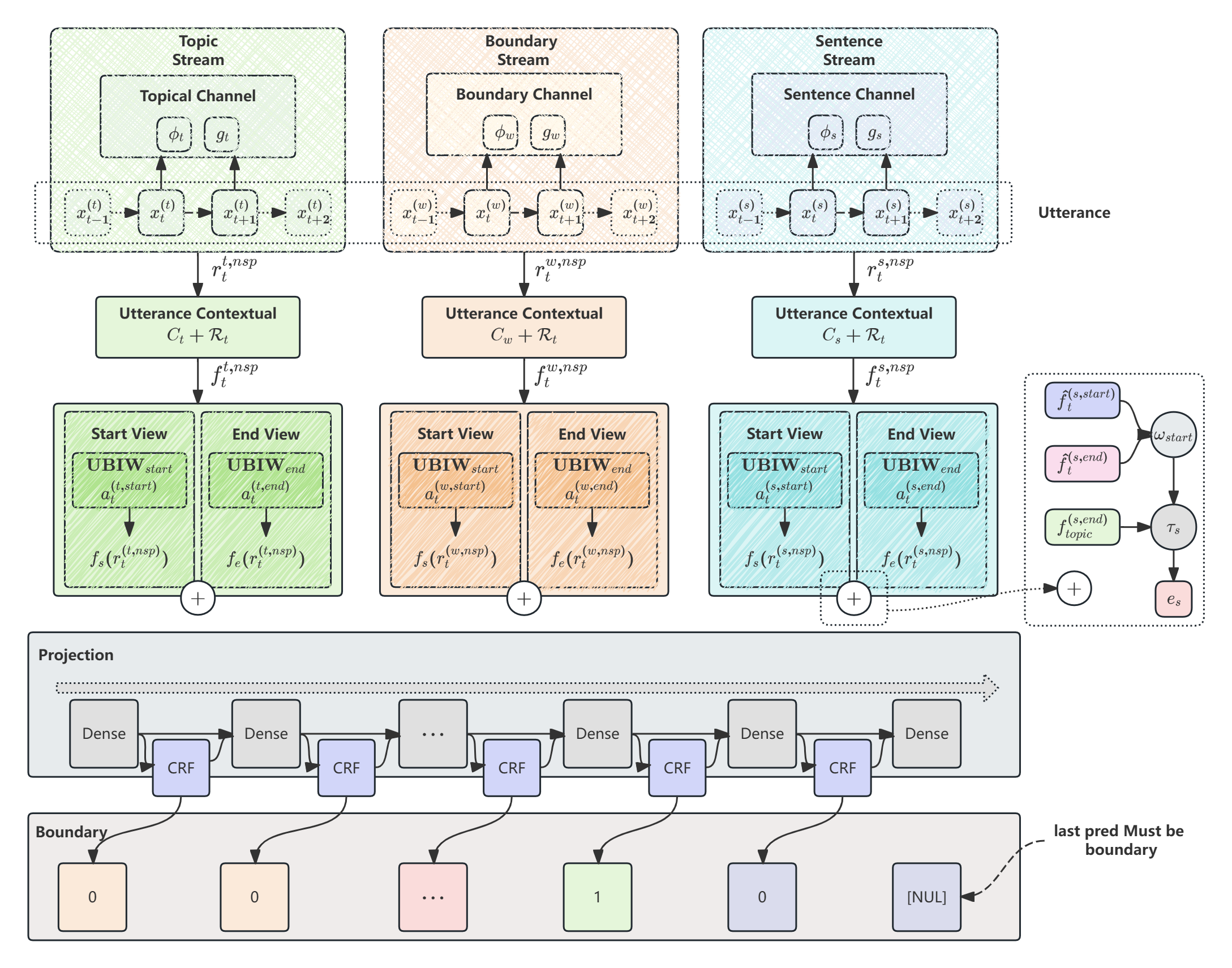}
    \caption{Detailed architecture of CobSeg.}
    \label{fig:main_structure}
\end{figure*}

Given a dialogue $\mathcal{D}=\{u_1,\ldots,u_T\}$ with $T$ utterances, dialogue topic segmentation predicts whether each adjacent utterance pair forms a topic boundary. The boundary sequence is defined as $\mathbf{y}=\{y_1,\ldots,y_{T-1}\}$, where $y_t=1$ indicates a topic boundary between $u_t$ and $u_{t+1}$. Thus, each prediction is associated with a cut position $t\in\{1,\ldots,T-1\}$ rather than an individual utterance. CobSeg performs structured boundary prediction over all candidate cut positions using a linear-chain CRF:
\begin{equation}
\mathbf{y}^{\ast}
=
\arg\max_{\mathbf{y}\in\{0,1\}^{T-1}}
\left[
\sum_{t=1}^{T-1} s_t(y_t;\mathcal{D})
+\sum_{t=1}^{T-2} A_{y_t,y_{t+1}}
\right],
\label{eq:task}
\end{equation}
where $s_t(y_t;\mathcal{D})$ is the local emission score at cut position $t$, and $A\in\mathbb{R}^{2\times 2}$ is the learned transition matrix.

\Cref{fig:main_structure} illustrates the overall architecture of CobSeg. The model computes each cut-level emission score from three complementary views of the dialogue: a Coherence Encoder that models semantic continuity across utterances, a Lexical Boundary Detector that preserves token-level transition cues near utterance edges, and a Topic Structure Extractor that provides corpus-informed topic coherence evidence. The same architecture is used in both supervised and pseudo-label settings; only the source of boundary supervision differs. Training minimizes the CRF negative log-likelihood over cut positions, augmented with lightweight auxiliary losses that encourage sparse UBIW scores and margin-based separation of boundary-adjacent tokens \citep{koshorek2018text}; full hyperparameters are listed in \Cref{tab:params}.

\subsection{Multi-View Transition Representation}

For an utterance $u_t$, let $\mathbf{x}_t^{(s)}\in\mathbb{R}^{d}$ denote the Coherence Encoder representation obtained by mean pooling the hidden states of the main encoder \citep{devlin2019bert,liu2019roberta}. Let $\mathbf{v}_{t,j}\in\mathbb{R}^{d}$ denote the contextual state of the $j$-th token in $u_t$, $\mathbf{p}_j$ its positional embedding, $m_{t,j}\in\{0,1\}$ the token mask, and $\rho_{t,j}\in[0,1]$ the normalized token position. Following the TextSeg backbone \citep{koshorek2018text}, the Lexical Boundary Detector applies a temporally sensitive structure for token sequence modeling within the utterance. Bidirectional context modeling allows the encoder to capture dependencies from both past and future tokens \citep{schuster1997bidirectional}. Let $\mathrm{Enc}_{\mathrm{tok}}$ denote a bidirectional LSTM encoder that processes token sequences within each utterance:
\begin{equation}
\mathbf{h}_{t,j}=\mathrm{Enc}_{\mathrm{tok}}(\mathbf{v}_{t,j}+\mathbf{p}_j),
\label{eq:token-enc}
\end{equation}
The resulting token states are pooled with an edge aware weighting rule:
\begin{equation}
\begin{split}
\pi_{t,j} &= \alpha + (1-\alpha)\lvert 2\rho_{t,j}-1\rvert^{\gamma}, \\
\mathbf{x}_t^{(w)} &=
\mathbf{W}_w
\frac{\sum_j m_{t,j}\pi_{t,j}\mathbf{h}_{t,j}}
{\sum_j m_{t,j}\pi_{t,j}},
\end{split}
\label{eq:token-pool}
\end{equation}
which upweights tokens near utterance boundaries. For the Topic Structure Extractor, $\mathbf{x}_t^{(t)}\in\mathbb{R}^{d}$ denotes a static topic representation obtained by mean-pooling an encoder oriented toward coherence \citep{gao2021simcse} over $u_t$.

For each branch $b\in\{s,w,t\}$, CobSeg converts utterance representations into cut oriented transition features. At cut position $t$, the left and right utterance states are compared through
\begin{equation}
\mathbf{z}_t^{(b)}=
\left[
\mathbf{x}_t^{(b)};
\mathbf{x}_{t+1}^{(b)};
\lvert \mathbf{x}_{t+1}^{(b)}-\mathbf{x}_t^{(b)}\rvert;
\mathbf{x}_t^{(b)}\odot \mathbf{x}_{t+1}^{(b)}
\right],
\label{eq:pair-feat}
\end{equation}
which summarizes persistence, change magnitude, and feature interaction. The branch state is then updated by two asymmetric gated residual adapters that share the same transition map $\phi_b$, scalar gate $g_b$, and transition feature $\mathbf{z}_t^{(b)}$, but anchor to opposite sides of the cut:
\begin{equation}
\begin{split}
\mathbf{r}_t^{(b,\mathrm{end})} &=
(1-\sigma(g_b))\mathbf{x}_t^{(b)}
+\sigma(g_b)\,\phi_b(\mathbf{z}_t^{(b)}), \\[4pt]
\mathbf{r}_t^{(b,\mathrm{start})} &=
(1-\sigma(g_b))\mathbf{x}_{t+1}^{(b)}
+\sigma(g_b)\,\phi_b(\mathbf{z}_t^{(b)}),
\end{split}
\label{eq:pair-fuse}
\end{equation}
where $\mathbf{r}_t^{(b,\mathrm{end})}$ anchors to the left utterance $u_t$ for modeling topic conclusion, and $\mathbf{r}_t^{(b,\mathrm{start})}$ anchors to the right utterance $u_{t+1}$ for modeling topic initiation. The learned scalar gate $g_b$ controls how much the transition correction replaces the base utterance representation in each direction.

The same NSP fusion and contextualization steps are applied to both directional adapter outputs. For brevity, $\mathbf{r}_t^{(b)}$ stands for either $\mathbf{r}_t^{(b,\mathrm{end})}$ or $\mathbf{r}_t^{(b,\mathrm{start})}$ in the equations that follow. The NSP channel is disabled in the supervised main results reported in \Cref{tab:sup-results}; in those experiments $\mathbf{r}_t^{(b,\mathrm{nsp})}=\mathbf{r}_t^{(b)}$ and $\beta_{\mathrm{nsp}}=0$. The NSP channel is enabled only in the pseudo-label setting, where the cross-encoder boundary probability $q_t$ provides an additional coherence signal to compensate for noisier training targets. When the optional NSP channel is enabled, adjacent utterance pairs are encoded by a cross encoder to produce an auxiliary pair representation $\mathbf{n}_t$ and a pairwise boundary probability $q_t$ \citep{devlin2019bert}. The auxiliary representation is injected into each branch as
\begin{equation}
\mathbf{r}_t^{(b,\mathrm{nsp})}=
\mathbf{r}_t^{(b)}+\sigma(\lambda_b)\mathbf{n}_t,
\label{eq:nsp-fuse}
\end{equation}
where $\lambda_b$ is a fusion gate specific to each branch. When the NSP channel is absent, $\mathbf{r}_t^{(b,\mathrm{nsp})}=\mathbf{r}_t^{(b)}$. Each branch is then contextualized by a temporally sensitive structure for sequence modeling at the dialogue level, following the TextSeg architecture \citep{koshorek2018text}. The contextualizer aggregates information across the dialogue sequence, allowing each cut position to access both preceding and following context \citep{schuster1997bidirectional,vaswani2017attention}:
\begin{equation}
\mathbf{f}_t^{(b)}=
\mathcal{C}_b(\mathbf{r}_{1:T-1}^{(b,\mathrm{nsp})},t)
+\mathbf{W}_b\mathbf{r}_t^{(b,\mathrm{nsp})},
\label{eq:context}
\end{equation}
where $\mathcal{C}_b$ is a temporally sensitive sequence encoder stacked over the transition features and the residual projection $\mathbf{W}_b$ preserves the local cut position signal after dialogue level encoding.

\subsection{Utterance Boundary Informativeness Weighting}

In a typical multi turn dialogue, only a small fraction of utterances carry decisive evidence about topic transitions. Most turns continue an ongoing topic without introducing lexical shifts. Treating every utterance position as equally informative dilutes the model's representational budget.

CobSeg addresses this through utterance boundary informativeness weighting, which learns to score each cut position by its predicted utility for boundary prediction and reweights the branch state accordingly. The module operates independently on the end view and start view of each branch. This allows the model to learn that different utterances may carry distinct informativeness for recognizing topic conclusions versus topic initiations. For instance, an utterance that summarizes a preceding topic is informative for the end direction, while one that introduces new terminology may be more informative for the start direction.

For branch state $\mathbf{f}_t^{(b)}$, each branch $b$ and direction $\star\in\{\mathrm{end},\mathrm{start}\}$ has its own scorer $\psi_{b,\star}$ and its own informativeness score $a_t^{(b,\star)}$. Let $\tilde{a}_t^{(b,\star)}$ denote the centered version over valid positions, and $\hat{\mathbf{f}}_t^{(b,\star)}$ the reweighted branch state. The reweighting mechanism draws inspiration from attention based feature weighting, where importance scores modulate the contribution of different positions \citep{bahdanau2014neural}. The reweighting takes the form
\begin{equation}
\begin{split}
a_t^{(b,\star)} &= \sigma(\psi_{b,\star}(\mathbf{f}_t^{(b)})), \\
\tilde{a}_t^{(b,\star)} &=
a_t^{(b,\star)}-
\frac{1}{T-1}\sum_{\tau=1}^{T-1} a_{\tau}^{(b,\star)}, \\
\hat{\mathbf{f}}_t^{(b,\star)} &=
\mathbf{f}_t^{(b)}
\left(1+\sigma(\xi)\tilde{a}_t^{(b,\star)}\right),
\end{split}
\label{eq:ubiw}
\end{equation}
where $\psi_{b,\star}$ is a lightweight informativeness scorer specific to branch $b$ and direction $\star$, and $\xi$ controls the residual scaling strength. The centering step makes the module redistributive rather than uniformly amplifying. Positions that appear more boundary informative are emphasized relative to less informative ones, while the mean activation across the dialogue is preserved. This design prevents the model from simply boosting all positions and forces it to make comparative judgments about which utterances genuinely contribute to boundary evidence. The reweighted states feed directly into the directional boundary heads described next.

\subsection{Directional Evidence Composition}

A topic boundary should indicate both the conclusion of the preceding topic and the initiation of a new one. These two signals are asymmetric: conclusion may appear through summary cues or reduced lexical specificity, while initiation may involve new terms, topic-setting phrases, or style shifts. CobSeg models this distinction with direction-specific prediction heads for the end view and start view. Both views share contextualized features but use separate UBIW scorers and classifiers, enabling lightweight directional specialization without separate encoders.

Given the reweighted branch states $\hat{\mathbf{f}}_t^{(b,\mathrm{end})}$ and $\hat{\mathbf{f}}_t^{(b,\mathrm{start})}$, each branch produces two-dimensional logits:
\begin{equation}
\begin{split}
\mathbf{e}_t^{(b,\mathrm{end})} &=
\mathrm{Head}_{b,\mathrm{end}}(\hat{\mathbf{f}}_t^{(b,\mathrm{end})}), \\
\mathbf{e}_t^{(b,\mathrm{start})} &=
\mathrm{Head}_{b,\mathrm{start}}(\hat{\mathbf{f}}_t^{(b,\mathrm{start})}),
\end{split}
\label{eq:heads}
\end{equation}
where the last component denotes the positive boundary class. Each head is a single linear projection, keeping per-branch overhead small.

The Coherence Encoder and Lexical Boundary Detector form the main evidence channel, combining semantic continuity and vocabulary-change signals. The Topic Structure Extractor is added as a residual expert:
\begin{equation}
\begin{split}
\mathbf{e}_t^{(\mathrm{main},\star)} &=
\omega\,\mathbf{e}_t^{(s,\star)}+
(1-\omega)\mathbf{e}_t^{(w,\star)}, \\
\mathbf{e}_t^{(\star)} &=
(1-\tau)\mathbf{e}_t^{(\mathrm{main},\star)}+
\tau\,\mathbf{e}_t^{(t,\star)},
\end{split}
\label{eq:stream-merge}
\end{equation}
with learned gates $\omega,\tau\in(0,1)$. This lets topic-level structure contribute when useful while preserving local lexical evidence. The positive component of $\mathbf{e}_t^{(\star)}$ is denoted $\bar{\ell}_{t,1}^{(\star)}$.

The final positive boundary logit further incorporates topic coherence cues and a class-prior bias:
\begin{equation}
\begin{split}
\ell_{t,1}^{(\star)}
= \bar{\ell}_{t,1}^{(\star)}
+ \beta_{\mathrm{coh}} k_t^{\mathrm{coh}}
+ \beta_{\mathrm{bnd}} k_t^{\mathrm{bnd}} \\
\quad
+ \beta_{\mathrm{nsp}}
\log \frac{q_t}{1-q_t}
+ \log w_{+},
\end{split}
\label{eq:bias}
\end{equation}
where $k_t^{\mathrm{coh}}$ and $k_t^{\mathrm{bnd}}$ are keyword-based topic coherence channels. The fixed bias $\log w_{+}$ compensates for the rarity of boundaries and is applied during both training and inference; $w_{+}=2.0$ is selected based on validation performance.

\begin{table*}[!h]
\centering
\caption{Supervised results for trainable DTS segmenters}
\label{tab:sup-results}
\resizebox{\textwidth}{!}{
\begin{tabular}{llccccccccccccccc}
\toprule
\multirow{2}{*}{Model} & \multirow{2}{*}{Source} & \multicolumn{3}{c}{VHF} & \multicolumn{3}{c}{DialSeg711} & \multicolumn{3}{c}{Doc2Dial} & \multicolumn{3}{c}{TIAGE} & \multicolumn{3}{c}{SuperSeg} \\
\cmidrule(lr){3-5}\cmidrule(lr){6-8}\cmidrule(lr){9-11}\cmidrule(lr){12-14}\cmidrule(lr){15-17}
& & $P_k\downarrow$ & $W_d\downarrow$ & $F_1\uparrow$ & $P_k\downarrow$ & $W_d\downarrow$ & $F_1\uparrow$ & $P_k\downarrow$ & $W_d\downarrow$ & $F_1\uparrow$ & $P_k\downarrow$ & $W_d\downarrow$ & $F_1\uparrow$ & $P_k\downarrow$ & $W_d\downarrow$ & $F_1\uparrow$ \\
\midrule
TextSeg & NAACL$^{2018}$ & 8.2 & 8.7 & 91.4 & 6.3 & 8.1 & 94.3 & 28.5 & 31.7 & 81.4 & 39.3 & 39.3 & 44.9 & 24.1 & 25.9 & \textbf{85.0} \\
BERT & NAACL$^{2019}$ & 7.8 & 8.7 & 88.9 & 2.9 & 3.8 & 94.8 & 28.1 & 30.3 & 71.3 & 31.5 & 33.8 & 52.7 & 23.5 & 24.8 & 77.7 \\
RoBERTa & arXiv$^{2019}$ & 4.3 & 4.8 & \underline{96.9} & 1.8 & 2.3 & 97.6 & 27.7 & \underline{29.6} & \textbf{86.7} & 32.2 & 33.7 & \underline{53.2} & 20.8 & 22.4 & 79.1 \\
TOD-BERT & EMNLP$^{2020}$ & 8.5 & 8.9 & 94.7 & \underline{1.1} & \underline{1.7} & \underline{98.0} & \underline{27.5} & \underline{29.6} & \underline{86.5} & 36.5 & 38.5 & 52.9 & \underline{20.0} & \underline{21.5} & \underline{79.6} \\
T5 & JMLR$^{2020}$ & 7.6 & 8.0 & 92.7 & 6.6 & 6.9 & 91.7 & 32.1 & 33.3 & 62.3 & 43.1 & 43.6 & 11.2 & 25.5 & 27.2 & 70.7 \\
RetroTS-T5 & EMNLP$^{2021}$ & 3.5 & 4.1 & 94.8 & 7.2 & 7.7 & 90.3 & 33.1 & 34.7 & 63.0 & 44.8 & 44.8 & 6.6 & 26.0 & 28.2 & 70.4 \\
FLAN-T5-Base & JMLR$^{2024}$ & 2.8 & 3.2 & 96.2 & 9.2 & 9.8 & 91.3 & 33.7 & 35.2 & 62.2 & 45.9 & 45.9 & 0.0 & 25.9 & 28.6 & 71.2 \\
~~~~-~Large & JMLR$^{2024}$ & \underline{2.6} & \underline{2.9} & 96.5 & 5.7 & 5.9 & 93.2 & 31.3 & 32.1 & 60.4 & \underline{31.1} & \underline{33.5} & 48.5 & 23.7 & 25.1 & 71.6 \\
SUPRP & NAACL$^{2025}$ & 3.1 & 3.7 & 96.6 & 1.5 & 2.4 & 97.3 & 44.6 & 46.8 & 38.1 & 31.3 & 37.5 & 46.9 & 27.1 & 23.9 & 77.3 \\
\midrule
\textbf{CobSeg} & (Ours) & \textbf{1.9} & \textbf{2.3} & \textbf{97.8} & \textbf{1.0} & \textbf{1.4} & \textbf{98.4} & \textbf{23.8} & \textbf{25.1} & 72.4 & \textbf{29.3} & \textbf{31.9} & \textbf{54.4} & \textbf{19.2} & \textbf{20.2} & \underline{80.3} \\
 &  & $\pm$ 0.01 & $\pm$ 0.03 & $\pm$ 0.99 & $\pm$ 0.02 & $\pm$ 0.02 & $\pm$ 0.81 & $\pm$ 0.28 & $\pm$ 0.33 & $\pm$ 1.15 & $\pm$ 0.51 & $\pm$ 0.27 & $\pm$ 0.43 & $\pm$ 0.46 & $\pm$ 0.51 & $\pm$ 1.13 \\
\bottomrule
\end{tabular}}
\par\smallskip\footnotesize$\downarrow$: Less is better;\quad $\uparrow$: More is better;\quad \textbf{BOLD}: $1^{st}$ per column;\quad \underline{UNDERLINE}: $2^{nd}$ per column;\quad $--$: not available.
\end{table*}

The keyword cues are built in two stages. First, training utterances are represented using semantic embeddings and TF-IDF features, then clustered into $K$ coarse topic regions by K-Means. For each region, TF-IDF identifies cohesive salient terms and ambient background terms. Boundary-related terms are extracted from utterances far from cluster centroids, while centroid-near terms serve as core reference vocabulary.

Second, during inference, each utterance is scored by keyword density without explicit region assignment. Let $\mathcal{C}_{\mathrm{sal}}$, $\mathcal{C}_{\mathrm{amb}}$, $\mathcal{B}_{\mathrm{mrk}}$, and $\mathcal{B}_{\mathrm{core}}$ denote the four keyword sets, and let
$\mathrm{dens}(u,\mathcal{S}) = |\mathrm{tokens}(u) \cap \mathcal{S}| / |\mathrm{tokens}(u)|$.
The coherence and boundary scores for cut position $t$ are:
\begin{equation}
\begin{aligned}
k_t^{\mathrm{coh}}
&= \sum_{i=t}^{t+1}
\Big[
\mathrm{dens}(u_i,\mathcal{C}_{\mathrm{sal}})
+ \lambda_{\mathrm{amb}}\mathrm{dens}(u_i,\mathcal{C}_{\mathrm{amb}})
\Big], \\
k_t^{\mathrm{bnd}}
&= \sum_{i=t}^{t+1}
\Big[
\mathrm{dens}(u_i,\mathcal{B}_{\mathrm{mrk}})
- \lambda_{\mathrm{core}}\mathrm{dens}(u_i,\mathcal{B}_{\mathrm{core}})
\Big].
\end{aligned}
\label{eq:kw-scores}
\end{equation}

with $\lambda_{\mathrm{amb}}=0.25$, $\lambda_{\mathrm{core}}=0.15$, $K=8$, and a salient keyword budget of 220 terms per dataset. These keyword sets are fixed after preprocessing; only $\beta_{\mathrm{coh}}$, $\beta_{\mathrm{bnd}}$, and $\beta_{\mathrm{nsp}}$ are learned. Since the coefficients can shrink toward zero, uninformative keyword cues can be ignored automatically. The ablation in \Cref{tab:ablation-main} shows that removing the coherence cue worsens $P_k$ and $W_d$, indicating that the fixed keyword partition provides useful signal.

The final cut emission averages end-view and start-view evidence:
\begin{equation}
\mathbf{e}_t^{\mathrm{cut}}=
\frac{1}{2}\mathbf{e}_t^{(\mathrm{end})}
+\frac{1}{2}\mathbf{e}_t^{(\mathrm{start})},
\label{eq:cut}
\end{equation}
for $t\in\{1,\ldots,T-1\}$. Here $\mathbf{e}_t^{(\mathrm{end})}$ scores topic conclusion from $u_t$, while $\mathbf{e}_t^{(\mathrm{start})}$ scores topic initiation from $u_{t+1}$. The CRF local score is $s_t(y_t;\mathcal{D})=[\mathbf{e}_t^{\mathrm{cut}}]_{y_t}$. Thus, a strong boundary requires evidence from both sides of the cut.

Training uses the same CRF decoding interface in both supervised and pseudo-label settings. The model minimizes negative log likelihood over the $T-1$ cut positions; only the source of boundary labels differs.

\section{Experiments}

\begin{table*}[!h]
\centering
\caption{Regime-aware DTS comparison under non-gold-boundary settings}
\label{tab:lf-results}
\resizebox{\textwidth}{!}{
\begin{tabular}{llccccccccccccccc}
\toprule
\multirow{2}{*}{Model} & \multirow{2}{*}{Source} & \multicolumn{3}{c}{VHF} & \multicolumn{3}{c}{DialSeg711} & \multicolumn{3}{c}{Doc2Dial} & \multicolumn{3}{c}{TIAGE} & \multicolumn{3}{c}{SuperSeg} \\
\cmidrule(lr){3-5}\cmidrule(lr){6-8}\cmidrule(lr){9-11}\cmidrule(lr){12-14}\cmidrule(lr){15-17}
& & $P_k\downarrow$ & $W_d\downarrow$ & $F_1\uparrow$ & $P_k\downarrow$ & $W_d\downarrow$ & $F_1\uparrow$ & $P_k\downarrow$ & $W_d\downarrow$ & $F_1\uparrow$ & $P_k\downarrow$ & $W_d\downarrow$ & $F_1\uparrow$ & $P_k\downarrow$ & $W_d\downarrow$ & $F_1\uparrow$ \\
\midrule
\multicolumn{17}{c}{\small\bf Classical Unsupervised} \\
\midrule
Random & -- & 50.4 & 78.4 & 37.4 & 53.4 & 71.5 & 40.7 & 53.0 & 67.8 & 42.4 & 51.8 & 66.1 & 42.8 & 49.5 & 65.4 & 45.1 \\
Even & -- & 50.7 & 74.7 & 38.6 & 49.3 & 69.3 & 44.4 & 52.6 & 67.2 & 44.1 & 50.8 & 66.6 & 42.2 & 50.6 & 68.4 & 44.3 \\
TeT & CL$^{1997}$ & 43.8 & 55.5 & 65.4 & 42.1 & 44.4 & 65.5 & 48.5 & 48.7 & \underline{68.0} & 44.2 & 44.9 & \underline{64.2} & 44.0 & 99.0 & 25.1 \\
TeT$_{\text{Glove}}$ & CL$^{2016}$ & 35.0 & 41.6 & \underline{72.0} & 44.0 & 45.6 & 63.6 & 52.9 & 95.4 & 25.7 & 46.8 & 47.5 & 63.9 & 44.0 & 99.1 & 25.0 \\
TeT$_{\text{CLS}}$ & CL$^{2021}$ & 41.1 & 47.1 & 67.0 & 37.0 & 38.9 & 66.4 & 53.3 & 85.3 & 38.7 & 44.3 & 44.9 & 63.8 & 43.9 & 98.9 & 25.1 \\
TeT$_{\text{NSP}}$ & CL$^{2021}$ & 35.6 & 39.1 & 25.2 & 43.4 & 44.1 & 15.1 & 50.6 & 50.6 & 12.0 & 48.6 & 49.1 & 10.5 & 46.3 & 97.9 & 41.6 \\
BayesSeg & EMNLP$^{2008}$ & 36.5 & 45.4 & 63.9 & 38.1 & 48.6 & 63.1 & 48.1 & 66.6 & 51.9 & 54.9 & 70.7 & 48.3 & 41.8 & 63.1 & \underline{55.6} \\
\midrule
\multicolumn{17}{c}{\small\bf Neural Unsupervised} \\
\midrule
GraphSeg & *SEM$^{2016}$ & 48.7 & 70.6 & 56.8 & 42.3 & 53.9 & 64.1 & 58.0 & 61.6 & 49.2 & 58.9 & 65.0 & 47.2 & 57.0 & 60.8 & 48.4 \\
T5 & JMLR$^{2020}$ & 46.9 & 46.9 & 0.0 & 43.5 & 43.7 & 2.3 & 47.7 & 48.0 & 3.9 & 46.3 & 47.2 & 8.4 & 52.9 & 53.2 & 4.7 \\
GreedySeg & AAAI$^{2021}$ & 41.5 & 49.5 & 55.8 & 38.6 & 43.0 & 66.4 & 48.6 & 49.0 & 59.7 & 49.1 & 50.4 & 49.1 & 54.5 & 55.0 & 41.1 \\
CSM & SIGDIAL$^{2021}$ & -- & -- & -- & 27.8 & 30.2 & 61.0 & -- & -- & -- & \underline{40.0} & \underline{42.0} & 42.7 & 46.2 & 46.7 & 38.1 \\
DynamicCOCO & NLPIR$^{2023}$ & -- & -- & -- & 18.4 & 21.1 & \textbf{83.2} & 42.0 & 45.0 & \textbf{70.1} & -- & -- & -- & -- & -- & -- \\
UMLF & TASLP$^{2024}$ & -- & -- & -- & \underline{15.8} & \underline{14.1} & -- & \underline{37.3} & 40.8 & -- & 40.7 & 44.4 & -- & -- & -- & -- \\
UUPRP & NAACL$^{2025}$ & 48.6 & 85.2 & 3.2 & 34.6 & 56.9 & 0.3 & 38.1 & \underline{39.5} & 0.3 & 50.3 & 50.6 & 4.7 & \underline{39.5} & \textbf{40.0} & 0.4 \\
\midrule
\multicolumn{17}{c}{\small\bf Unsupervised with Induced Training Signals} \\
\midrule
DialSTART & SIGIR$^{2023}$ & \underline{25.4} & \underline{26.4} & 49.2 & 28.0 & 31.7 & 48.3 & 44.1 & 48.9 & 46.9 & 43.8 & 49.9 & 38.3 & \textbf{39.1} & 45.4 & 48.5 \\
DyDTS & WWW$^{2025}$ & 35.3 & 37.9 & 37.8 & 26.1 & 30.0 & 61.6 & 51.3 & 52.7 & 29.5 & 44.9 & 48.3 & 35.6 & 49.7 & 51.2 & 31.5 \\
SumSeg & NAACL$^{2024}$ & 37.6 & 37.7 & 60.1 & 33.6 & 34.4 & 59.5 & 41.9 & 42.2 & 60.6 & 41.0 & 42.2 & 58.1 & 49.6 & 50.0 & \textbf{57.0} \\
\midrule
\multicolumn{17}{c}{\small\bf LLM Direct Inference} \\
\midrule
S3-DST & ACL$^{2024}$ & -- & -- & -- & 8.7 & 10.9 & 79.0 & -- & -- & -- & 43.9 & 49.8 & 26.5 & 44.2 & 46.9 & 40.4 \\
Def-DTS & ACL$^{2025}$ & -- & -- & -- & 1.5 & 1.8 & 97.9 & -- & -- & -- & 23.2 & 25.6 & 69.9 & 31.5 & 32.4 & 68.6 \\
\midrule
\textbf{CobSeg} & (Ours) & \textbf{10.6} & \textbf{11.4} & \textbf{86.2} & \textbf{14.3} & \textbf{13.8} & \underline{75.5} & \textbf{35.6} & \textbf{38.7} & 62.4 & \textbf{38.9} & \textbf{40.2} & \textbf{65.7} & 40.7 & \underline{42.5} & 52.4 \\
\bottomrule
\end{tabular}}
\end{table*}

\subsection{Setup}

We distinguish trainable segmentation architectures from direct LLM inference methods in our evaluation scope. The supervised comparison covers trainable segmenters on five benchmarks: VHF \citep{sun2025dash}, DialSeg711 \citep{xu2021topic}, Doc2Dial \citep{feng2020doc2dial}, TIAGE \citep{xie2021tiage}, and SuperSeg \citep{jiang2023superdialseg}. Evaluation uses $P_k$ \citep{beeferman1999statistical}, $W_d$ \citep{pevzner2002critique}, and boundary $F_1$, with all methods evaluated under identical train/validation/test splits in \Cref{tab:appendix-stats}, a unified metric implementation, and the same hyperparameter tuning protocol; hyperparameters are listed in \Cref{tab:params}. Direct LLM inference is excluded from the main comparison because it relies on large general-purpose models and prompt-based reasoning at inference time, whereas CobSeg performs boundary prediction with a learned segmenter and does not call an LLM during inference.

\subsection{Main Results}

\subsubsection{Supervised results for trainable DTS}

\Cref{tab:sup-results} reports the supervised comparison aamong trainable DTS segmenters. Baseline values are obtained under the split configuration and hyperparameter tuning protocol described in \Cref{sec:appendix-data}.

CobSeg achieves the lowest $P_k$ and $W_d$ on 5 of 5 datasets and the highest $F_1$ on 3 of 5 datasets. The remaining two datasets show a consistent trade off: CobSeg improves local cut positioning but does not always produce the best boundary-rate calibration. On Doc2Dial and SuperSeg, encoder baselines obtain higher $F_1$ despite worse $P_k$ and $W_d$, indicating that sharper local boundary decisions do not fully determine the global number of emitted boundaries.

\subsubsection{Regime-aware comparison under NGB}

\Cref{tab:lf-results} reports results under None-Gold-Boundary(NGB) settings, covering classical and neural unsupervised methods, methods trained with induced supervision, direct LLM inference methods, and CobSeg pseudo-label. Because these methods differ in how boundary signals are obtained and how inference is performed, the table should be interpreted as a regime-aware comparison rather than a single unified leaderboard. In particular, direct LLM inference methods perform prompt-based reasoning at test time, whereas CobSeg pseudo-label trains a learned segmenter from induced supervision and performs inference without LLM calls.

Overall, CobSeg pseudo-label achieves the strongest performance among inference-free methods on most datasets. It obtains the best $P_k$ and $W_d$ on VHF, DialSeg711, Doc2Dial, and TIAGE, with especially large gains on VHF. It also achieves the best $F_1$ on VHF and TIAGE, and remains competitive on DialSeg711 and Doc2Dial, although DynamicCOCO gives higher $F_1$ on these two datasets. On SuperSeg, CobSeg pseudo-label is competitive but not dominant: it improves over most unsupervised baselines in $P_k$ and $W_d$, but SumSeg achieves the best $F_1$ and Def-DTS remains stronger under direct LLM inference. These results suggest that pseudo-label training can substantially narrow the gap to stronger supervision, while its effectiveness still depends on the quality of induced pseudo segments, especially on larger and more open-domain corpora.

\subsection{Pseudo Label Source Analysis}
\label{sec:pseudo-source}

To compare pseudo-label quality, four TeT-based boundary generators are evaluated directly against gold test annotations on DialSeg711 and Doc2Dial. They share the same peak-picking backbone and differ only in the adjacent-utterance similarity signal: TF-IDF, GloVe, BERT CLS, or BERT NSP. No segmenter is trained in this analysis; the goal is to measure induction quality alone. Reconstructed pseudo-boundaries from the CobSeg pipeline are evaluated under the same protocol and are comparable to these TeT-based sources.

\begin{table}[!htbp]
\centering
\small
\caption{Pseudo-label source quality on DialSeg711 and Doc2Dial, evaluated at the 100-dialogue scale.}
\label{tab:pseudo-source}
\resizebox{\columnwidth}{!}{
\begin{tabular}{lcccccc}
\toprule
\multirow{2}{*}{Method} & \multicolumn{3}{c}{DialSeg711} & \multicolumn{3}{c}{Doc2Dial} \\
\cmidrule(lr){2-4}\cmidrule(lr){5-7}
 & $P_k\downarrow$ & $W_d\downarrow$ & $F_1\uparrow$ & $P_k\downarrow$ & $W_d\downarrow$ & $F_1\uparrow$ \\
\midrule
TeT                     & 42.0 & 44.0 & 19.0 & 48.0 & 49.0 &  7.0 \\
~~+~Glove               & 33.6 & 39.7 & 16.2 & 53.1 & 91.2 & 32.9 \\
~~+~CLS                 & 40.8 & 46.0 & 64.2 & 54.7 & 92.6 & 36.4 \\
~~+~NSP                 & 35.2 & 28.4 & 31.8 & 51.2 & 51.7 & 34.0 \\
\midrule
DialSTART & 32.7 & 34.1 & 42.3 & 47.5 & 52.4 & 42.6 \\
DyDTS     & 42.8 & 44.2 & 23.0 & 53.4 & 54.9 & 26.6 \\
\bottomrule
\end{tabular}}
\end{table}

\begin{table*}[!h]
\centering
\tiny
\setlength{\tabcolsep}{3.5pt}
\caption{Full supervision sweep on the test split.}
\label{tab:budget}
\resizebox{\textwidth}{!}{
\begin{tabular}{lcccccccccccccccccccc}
\toprule
\multirow{2}{*}{Train \%} & \multicolumn{4}{c}{VHF} & \multicolumn{4}{c}{DialSeg711} & \multicolumn{4}{c}{Doc2Dial} & \multicolumn{4}{c}{TIAGE} & \multicolumn{4}{c}{SuperSeg} \\
\cmidrule(lr){2-5}\cmidrule(lr){6-9}\cmidrule(lr){10-13}\cmidrule(lr){14-17}\cmidrule(lr){18-21}
 & $P_k\downarrow$ & $W_d\downarrow$ & $F_1\uparrow$ & $\Delta P_k$ & $P_k\downarrow$ & $W_d\downarrow$ & $F_1\uparrow$ & $\Delta P_k$ & $P_k\downarrow$ & $W_d\downarrow$ & $F_1\uparrow$ & $\Delta P_k$ & $P_k\downarrow$ & $W_d\downarrow$ & $F_1\uparrow$ & $\Delta P_k$ & $P_k\downarrow$ & $W_d\downarrow$ & $F_1\uparrow$ & $\Delta P_k$ \\
\midrule
1\% & 46.9 & 46.9 & 0.0 & +45.0 & 43.3 & 43.3 & 0.0 & +42.3 & 29.8 & 31.6 & 67.0 & +6.0 & 45.9 & 45.9 & 0.0 & +16.6 & 26.7 & 28.1 & 75.6 & +7.5 \\
3\% & 46.9 & 46.9 & 0.0 & +45.0 & 3.7 & 4.5 & 94.5 & +2.7 & 28.7 & 30.5 & 69.5 & +4.9 & 45.0 & 46.4 & 20.8 & +15.7 & 26.2 & 27.6 & 74.8 & +7.0 \\
5\% & 9.9 & 12.1 & 84.2 & +8.0 & 4.0 & 5.1 & 94.2 & +3.0 & 28.3 & 29.9 & 68.7 & +4.5 & 40.6 & 43.0 & 41.1 & +11.3 & 25.6 & 27.1 & 76.3 & +6.4 \\
10\% & 9.9 & 12.2 & 84.6 & +8.0 & 2.9 & 3.6 & 95.5 & +1.9 & 28.4 & 30.3 & 68.5 & +4.6 & 41.0 & 44.1 & 38.7 & +11.7 & 24.9 & 26.1 & 74.3 & +5.7 \\
25\% & 5.9 & 6.8 & 91.1 & +3.9 & 1.7 & 2.2 & 97.5 & +0.7 & 26.7 & 28.6 & 71.4 & +2.9 & 31.1 & 32.3 & 46.5 & +1.8 & 24.0 & 25.4 & 75.8 & +4.8 \\
50\% & 7.6 & 9.1 & 88.7 & +5.7 & 1.3 & 1.6 & 97.7 & +0.3 & 25.7 & 27.0 & 70.0 & +1.9 & 30.8 & 33.1 & 50.8 & +1.5 & 22.7 & 23.8 & 77.6 & +3.5 \\
75\% & 4.9 & 6.0 & 93.0 & +3.0 & 1.6 & 2.1 & 97.7 & +0.6 & 25.5 & 26.6 & 69.0 & +1.7 & 29.9 & 31.8 & 52.7 & +0.6 & 22.3 & 23.4 & 76.9 & +3.1 \\
\midrule
100\% & 1.9 & 2.3 & 97.8 & -- & 1.0 & 1.4 & 98.4 & -- & 23.8 & 25.1 & 72.4 & -- & 29.3 & 31.9 & 54.4 & -- & 19.2 & 20.2 & 80.3 & -- \\
\bottomrule
\end{tabular}}
\end{table*}

\Cref{tab:pseudo-source} shows that pseudo-label quality is uneven across metrics. TeT$_{\text{NSP}}$ gives the best $P_k$/$W_d$ trade off among TeT variants, while TeT$_{\text{CLS}}$ tends to maximize $F_1$ by emitting more boundaries. DialSTART is competitive under the same 100-dialogue protocol, but it does not clearly dominate the TeT-based sources for producing training boundaries. Therefore, the stronger downstream results of CobSeg pseudo-label in \Cref{tab:lf-results} should be attributed primarily to the segmenter architecture rather than to unusually strong pseudo-label induction.

\subsection{Label Budget Analysis}

To measure label efficiency, the supervised model is trained with progressively smaller fractions of the original training set. \Cref{tab:budget} reports the sweep and uses the 100\% supervised result as the reference point. VHF and DialSeg711 approach full-data performance quickly, while Doc2Dial, TIAGE, and SuperSeg improve more gradually and require broader supervision for stable boundary localization.

\subsection{Ablation Study}

To isolate the contribution of the main design choices, the ablation study keeps the training setting, optimization recipe, and backbone encoder fixed while removing one component at a time. The four ablations remove the Lexical Boundary Detector, boundary informativeness weighting, topic coherence cue, and directional boundary heads on Doc2Dial and SuperSeg.

\begin{table}[!htbp]
\centering
\small
\caption{Component ablations on Doc2Dial and SuperSeg.}
\label{tab:ablation-main}
\resizebox{\columnwidth}{!}{
\begin{tabular}{lcccccc}
\toprule
\multirow{2}{*}{Model} & \multicolumn{3}{c}{Doc2Dial} & \multicolumn{3}{c}{SuperSeg} \\
\cmidrule(lr){2-4}\cmidrule(lr){5-7}
 & $P_k\downarrow$ & $W_d\downarrow$ & $F_1\uparrow$ & $P_k\downarrow$ & $W_d\downarrow$ & $F_1\uparrow$ \\
\midrule
w/o Token stream & 25.0 & 26.4 & 70.6 & 20.9 & 22.1 & 78.2 \\
w/o Inform. weight & 24.2 & 25.2 & 69.6 & 20.6 & 21.9 & 77.6 \\
w/o Coherence cue & 24.6 & 26.1 & 69.3 & 20.7 & 22.1 & 78.4 \\
w/o Bidirectional & 25.5 & 27.0 & 71.7 & 20.4 & 21.7 & 78.9 \\
\midrule
CobSeg & 23.8 & 25.1 & 72.4 & 19.2 & 20.2 & 80.3 \\
\bottomrule
\end{tabular}}
\end{table}

\Cref{tab:ablation-main} shows that all components contribute to the final performance. Removing the topic coherence cue consistently worsens $P_k$ and $W_d$, suggesting that corpus-level statistical signals complement learned representations. The Lexical Boundary Detector and informativeness weighting bring dataset-dependent gains, with larger effects when topic shifts coincide with clear lexical changes near utterance boundaries. Collapsing the directional end/start heads into a single score also degrades $P_k$ and $W_d$, confirming the benefit of modeling topic conclusion and initiation separately.

\subsection{Discussion}

The gap between $P_k$/$W_d$ and $F_1$ on Doc2Dial and SuperSeg suggests a granularity-calibration trade-off \citep{coen2025f1}: CobSeg sharpens cut-level boundary localization but does not fully regulate the global boundary rate in long dialogues. Adaptive thresholding or corpus-level segment length priors may further improve this aspect.

Ablation results show that topic coherence provides the most consistent contribution, while the Lexical Boundary Detector, UBIW, and directional boundary heads contribute in dataset-dependent ways. This supports CobSeg's core hypothesis that dialogue topic shifts benefit from combining local lexical transition evidence, directional end/start signals, and corpus-level coherence cues, rather than relying only on utterance-level semantic representations.

The controlled comparison in \Cref{sec:pseudo-source} identifies pseudo-segment fidelity as the main bottleneck in the pseudo-label setting. Although CobSeg remains competitive among methods without test-time LLM calls, low-quality boundary induction limits its gains, suggesting iterative joint refinement of the inducer and segmenter as future work. More broadly, CobSeg should be viewed as a compact trainable boundary model rather than a replacement for direct LLM reasoning: unlike prompt-based LLM inference, it performs segmentation with a learned model and is most useful when efficiency, reproducibility, controllability, deployment cost, or privacy constraints matter.

\section{Conclusion}
CobSeg models topic shifts through token-level lexical cues, directional prediction heads, and informativeness weighting. Under gold supervision, it reduces $P_k$ by 0.7 points and $W_d$ by 0.6 points on VHF, while reaching $P_k$ of 1.0 on DialSeg711. With induced boundaries, the largest gain is a 14.8 points $P_k$ reduction on VHF; on DialSeg711 and TIAGE, $W_d$ drops by 0.3 and 1.8 points, respectively, outperforming all non‑LLM methods.

The strongest gains appear on VHF, DialSeg711, and TIAGE, where topic shifts are often marked by local lexical signals. 
The ablation study shows that the topic coherence cue provides the most consistent contribution, while lexical boundary modeling, informativeness weighting, and directional prediction further improve boundary localization in a dataset-dependent manner.

A gap remains between local cut placement and global calibration: CobSeg achieves lower $P_k$ and $W_d$ but trails some encoder baselines in boundary $F_1$ on Doc2Dial and SuperSeg. 
This indicates that the model is effective at sharpening individual boundary decisions, but still requires better control of boundary rate across longer dialogues. 
The pseudo-label experiments further show that training with automatically induced boundaries can recover substantial performance when gold annotations are unavailable, but its effectiveness depends on pseudo-segment quality.

Overall, the results suggest that explicit boundary-centered modeling remains valuable for dialogue topic segmentation, up to 14.8 $P_k$ and 15.0 $W_d$ imporved in induced VHF setting, particularly when efficient, reproducible, and controllable inference without test-time LLM calls is required. Progress in NGB segmentation will require joint advances in segmenter design and pseudo-label induction.

\section{Limitations}
Current study has several limitations. First, CobSeg still lags behind the best encoder baselines in boundary $F_1$ on several large or open-domain corpora, consistent with the granularity-sensitivity findings of \citet{coen2025f1}. While the model improves local boundary discrimination through directional fusion and coherence-aware logits, its decoding stage does not explicitly calibrate the number of emitted boundaries against corpus-level segment length distributions, which can lead to a mismatch between sharper cut-level decisions and global boundary-rate control; adaptive thresholding or corpus-level boundary-rate priors may help address this. Second, the pseudo-label setting depends on the quality of pseudo-segment induction. Since pseudo-segments provide a less stable supervision signal than manual annotations, noisy boundary induction can limit the effectiveness of the learned segmenter, and jointly optimizing the pseudo-segment inducer and the segmenter through iterative refinement is a natural direction for future work. Third, our results should not be interpreted as showing that compact trainable segmenters universally outperform direct LLM inference. Direct LLM methods and CobSeg operate under different inference regimes, with comparisons affected by model scale, prompting strategy, inference cost, and access to large general-purpose models. CobSeg instead targets a complementary setting where efficient, reproducible, and controllable segmentation without test-time LLM calls is desirable.

\section*{Acknowledgement}

This research was funded by \textbf{ABC} with funding grant number \textbf{ABC} by \textbf{ABC}.

\clearpage
\newpage

\appendix
\section{Related Work}

Early segmentation methods identify topic boundaries mainly through lexical cohesion. TextTiling locates topic shifts at valleys of lexical similarity \citep{hearst1997text}, BayesSeg models segmentation as a Bayesian generative process \citep{eisenstein2008bayesian}, and GraphSeg constructs semantic relatedness graphs to infer topical structure \citep{glavavs2016unsupervised}. These approaches established lexical and semantic discontinuity as key signals for segmentation, but they were largely designed for monologic texts rather than dialogues, where topic shifts are often implicit, interaction-driven, and context-dependent.

Supervised neural methods further advance dialogue segmentation by leveraging pretrained or task-specific encoders for utterance representation. Early work enhances TextTiling with GloVe embeddings \citep{song2016dialogue,pennington2014glove}, while subsequent studies introduce attention mechanisms \citep{badjatiya2018attention}, pointer networks \citep{li2018segbot}, and supervised hierarchical architectures such as TextSeg \citep{koshorek2018text}. With the rise of pretrained language models, BERT, RoBERTa, and TOD-BERT provide stronger contextualized representations for segmentation \citep{devlin2019bert,liu2019roberta,wu2020tod}. Building on these encoders, later methods stack transformer layers for sequence modeling \citep{lo2021transformer}, incorporate two-level transformer architectures with coherence modeling \citep{somasundaran2020two}, apply BERT to meeting segmentation \citep{solbiati2021unsupervised}, or integrate CRF layers for structured prediction \citep{nair2023neural}. Although these methods differ in their encoding and decoding strategies, they typically operate at the sentence or utterance level by compressing each unit into a fixed vector representation. This design may obscure fine-grained boundary cues, since boundary-adjacent tokens are averaged together with mid-utterance content, potentially diluting lexical transition signals.

Coherence-based methods model topic continuity through utterance pair scoring. Xing and Carenini propose coherence scoring, referred to as CSM \citep{xing2021improving}. \citep{xu2019cross} develop cross-domain coherence models. \citep{xia2022dialogue} use parallel extraction with neighbor smoothing. \citep{yang2025unified} unify supervised and unsupervised DTS through utterance pair modeling. \citep{xu2025unsupervised} jointly learn discourse parsing and segmentation. These methods focus on pairwise semantic coherence at the sentence level without explicit token transition modeling, which may miss lexical cues concentrated at utterance boundaries.

Topic modeling methods derive signals from latent structure. \citep{du2013topic} apply structured topic models. \citep{gong2022tipster} propose topic-guided language models. Liu et al. jointly optimize segmentation and topic categorization with shared representations \citep{liu2023joint}. \citep{park2023unsupervised} perform segmentation in hyperdimensional space. \citep{hou2024unsupervised} use utterance rewriting to support topic detection. \citep{artemiev2024leveraging} leverage summarization as weak supervision. \citep{inan2022structured} frame segmentation as generation task.

Topic shift and LLM methods address DTS through transition modeling. \citep{sevegnani2021otters} study one-turn transitions. \citep{yang2022take} propose topic shift aware knowledge selection. \citep{lin2023multi} use multi-granularity prompts. \citep{hwang2024mp2d} generate topic shift dialogues. \citep{takanobu2018weakly} apply reinforcement learning. \citep{vijjini2023curricular} use curricular pretraining. \citep{fan2024uncovering} explore ChatGPT for discourse analysis. Recent state-of-the-art methods include unified frameworks and LLM-based reasoning. \citep{yang2025unified} unify supervised and unsupervised DTS through utterance pair relation modeling, achieving strong performance through sentence pairing. \citep{lee2025defdts} cast DTS as deductive reasoning with LLMs, demonstrating effectiveness through structured inference. \citep{das2024s3} propose S3-DST for structured dialogue segmentation. \citep{lv2025dynamic} enhance boundaries with topic-aware propagation through dynamic mechanisms. CobSeg complements these approaches by explicitly modeling token lexical transitions, learning boundary informativeness weighting to concentrate capacity on decisive positions, and integrating corpus statistical cues with learned coefficients rather than as fixed preprocessing.

\section{Additional Dataset Statistics}
\label{sec:appendix-data}

\Cref{tab:appendix-stats} summarize the benchmark characteristics and provide additional distributional information used to contextualize the cross-dataset comparisons. The maximum sequence lengths in \Cref{tab:params} (48/48/64/32/64) are manually selected to cover approximately the 90th percentile of dialogue turns for each dataset, ensuring that most dialogues fit within the context window without excessive padding. The number of K-Means clusters is set to $K=8$, following prior work that partitions dialogue utterances into a small number of intent categories via LLM-based or clustering-based methods \citep{lee2025defdts,das2024s3,ma2024multi}. The salient keyword budget is set to 220 terms per dataset, selected as the point of diminishing TF-IDF cross-cluster distinctiveness gain in a sweep over candidate values.

\begin{table*}[!htbp]
\centering
\small
\caption{Dataset statistics computed from the benchmark files in the project repository.}
\label{tab:appendix-stats}
\resizebox{\textwidth}{!}{
\begin{tabular}{lrrrrrrrrr}
\toprule
Dataset & Train & Val & Test & Median Turns & 90th pct. Turns & Median Seg. Len. & 90th pct. Seg. Len. & Avg. Tokens/Turn & Boundaries/100 Turns \\
\midrule
VHF & 105 & 22 & 27 & 49.0 & 62.0 & 8.0 & 9.0 & 12.8 & 11.3 \\
DialSeg711 & 497 & 106 & 108 & 26.0 & 36.0 & 6.0 & 10.0 & 12.4 & 15.1 \\
Doc2Dial & 2289 & 490 & 491 & 13.0 & 15.0 & 2.0 & 7.0 & 14.5 & 19.1 \\
TIAGE & 300 & 100 & 100 & 16.0 & 16.0 & 3.0 & 7.0 & 10.4 & 19.4 \\
SuperSeg & 6948 & 1322 & 1322 & 12.0 & 15.0 & 2.0 & 5.0 & 14.4 & 25.8 \\
\bottomrule
\end{tabular}
}
\end{table*}

\section{Implementation Details}
\label{sec:appendix-impl}

The CobSeg architecture is optimized with the hyperparameter configuration in \Cref{tab:params}; the same settings are used for both supervised training and unsupervised pseudo-label training unless stated otherwise.

\begin{table}[!htbp]
\centering
\small
\caption{Key hyperparameters and training configuration of CobSeg.}
\label{tab:params}
\begin{tabular}{lr}
\toprule
\multicolumn{1}{c}{Parameter} & \multicolumn{1}{c}{Value} \\
\midrule
Coherence encoder & SimCSE \\
Primary encoder & RoBERTa-base \\
Contextualizer & 2 \\
$\alpha$ (edge bias) & 0.1 \\
$\gamma$ (sharpness) & 2.0 \\
$\xi$ (UBIW residual) & 0.3 \\
Optimizer & AdamW \\
LR & $2\times10^{-5}$ \\
Weight decay & 0.01 \\
Batch size & 16 \\
Max epochs & 50 \\
Early stop & 10 \\
Warmup ratio & 0.1 \\
Max seq len & 48/48/64/32/64 \\
Pos weight $w_{+}$ & 2.0 \\
$w_{+}$ candidates & $\{1.0,1.5,2.0,3.0,5.0\}$ \\
UBIW aux weight & 0.2 \\
UBIW decay $\tau$ & 2.0 \\
Lexical loss weight & 0.2 \\
Lexical margin & 0.1 \\
Lexical keyword gap & 0.05 \\
Stage-1 LR & $5\times10^{-4}$ \\
Stage-1 aux weight & 0.5 \\
NSP stage-2 aux weight & 0.2 \\
pseudo-label & deepseek-v4-pro \\
pseudo-label temp. & 0.1 \\
all rand. seed & 42 \\
top k & 1.0 \\
\bottomrule
\end{tabular}
\end{table}

\section{Computational Cost}
\label{sec:appendix-cost}

\Cref{tab:compute-cost} compares the parameter count, inference latency, and GPU memory footprint against representative supervised baselines. All measurements are obtained under the supervised setting with identical hardware and batch size. Parameter counts are in millions; GPU memory is in MB. CobSeg (architecture) refers to the modules proposed in this paper without any pretrained encoder. The intermediate rows show incremental additions: the Lexical Boundary Detector, UBIW module, and Topic Structure Extractor. CobSeg (full) adds two pretrained encoders (RoBERTa-base for Coherence Encoder / Lexical Boundary Detector and SimCSE for the Topic Structure Extractor) to the complete CobSeg architecture. The parameter overhead directly attributable to the proposed method is the gap between CobSeg (architecture) and a single-encoder baseline, approximately 24M parameters.

\begin{table}[!htbp]
\centering
\small
\caption{Computational cost comparison under the supervised setting.}
\label{tab:compute-cost}
\resizebox{\columnwidth}{!}{
\begin{tabular}{lrrr}
\toprule
Model & \#Param & ms/dial & GPU Mem \\
\midrule
RoBERTa  & 124.65 & 6.8 & 511 \\
BERT     & 109.48 & 7.1 & 454 \\
TOD-BERT & 109.48 & 7.1 & 454 \\
T5       & 109.63 & 9.0 & 457 \\
\midrule
CobSeg (full)               & 272.65 & 31.4 & 1131 \\
\ \ \ CobSeg (architecture)   & 23.88  & 7.6  & 236  \\
\ \ \ ~~+~ Lexical Boundary Detector       & 31.13  & 7.9  & 236  \\
\ \ \ ~~+~ UBIW               & 31.20  & 8.9  & 236  \\
\ \ \ ~~+~ Topic Structure Extractor      & 38.52  & 9.1  & 236  \\
\bottomrule
\end{tabular}}
\end{table}

\section{Baseline Descriptions}

Unless an official split is provided by the original benchmark release, dataset splits follow an approximate 7:1.5:1.5 train/validation/test ratio. For TIAGE, the official split configuration released by the original authors is used directly. For VHF, the split is derived from the official DASH dataset release. For SuperSeg, the official split released with the benchmark is used. For DialSeg711, the public dataset split is used. For Doc2Dial, the original dataset is publicly available; the segmentation split is constructed following the topic annotations provided in the official release. All experiments are conducted on these split settings. Where baseline values are reproduced under the splits used in this work, this is noted in the results.

\paragraph{Supervised Setting.}
CobSeg is compared with TextSeg \citep{koshorek2018text}, BERT \citep{devlin2019bert}, RoBERTa \citep{liu2019roberta}, TOD-BERT \citep{wu2020tod}, T5 \citep{raffel2020exploring}, FLAN-T5 \citep{chung2024scaling}, RetroTS-T5 \citep{xie2021tiage}, and SUPRP \citep{yang2025unified}. TextSeg uses hierarchical LSTMs with CRF decoding for structured boundary prediction \citep{nair2023neural}. BERT, RoBERTa, and TOD-BERT are pretrained transformer encoders fine-tuned for boundary classification. T5 and FLAN-T5 are encoder-decoder models adapted for segmentation as a sequence-to-sequence task. RetroTS-T5 extends T5 with retrospective topic shift detection.

\paragraph{Unsupervised Setting.}
CobSeg is compared with Random, Even, BayesSeg \citep{eisenstein2008bayesian}, TeT \citep{hearst1997text}, TeT$_{\text{Glove}}$ \citep{song2016dialogue,pennington2014glove}, TeT$_{\text{CLS}}$, TeT$_{\text{NSP}}$, GraphSeg \citep{glavavs2016unsupervised}, GreedySeg \citep{xu2021topic}, CSM \citep{xing2021improving}, T5 \citep{raffel2020exploring}, DialSTART \citep{gao2023unsupervised}, DynamicCOCO \citep{pu2023dialogue}, UMLF \citep{xu2025unsupervised}, SumSeg \citep{artemiev2024leveraging}, DyDTS \citep{lv2025dynamic}, UUPRP \citep{yang2025unified}, Def-DTS \citep{lee2025defdts}, and S3-DST \citep{das2024s3}. TeT$_{\text{Glove}}$ enhances TextTiling with GloVe embeddings \citep{song2016dialogue}. TeT$_{\text{CLS}}$ and TeT$_{\text{NSP}}$ replace lexical similarity with BERT-based sentence embeddings and next-sentence prediction scores, respectively. CSM models utterance-pair coherence. DialSTART learns topical representations through contrastive learning. SumSeg uses summarization as weak supervision. UMLF jointly learns discourse parsing and segmentation. DyDTS applies dynamic topic-aware propagation. UUPRP unifies utterance-pair relations. Def-DTS and S3-DST use LLMs for structured reasoning at inference time.

\section{Qualitative Analysis}

\begin{figure*}[!htbp]
\centering
\includegraphics[width= 2\columnwidth]{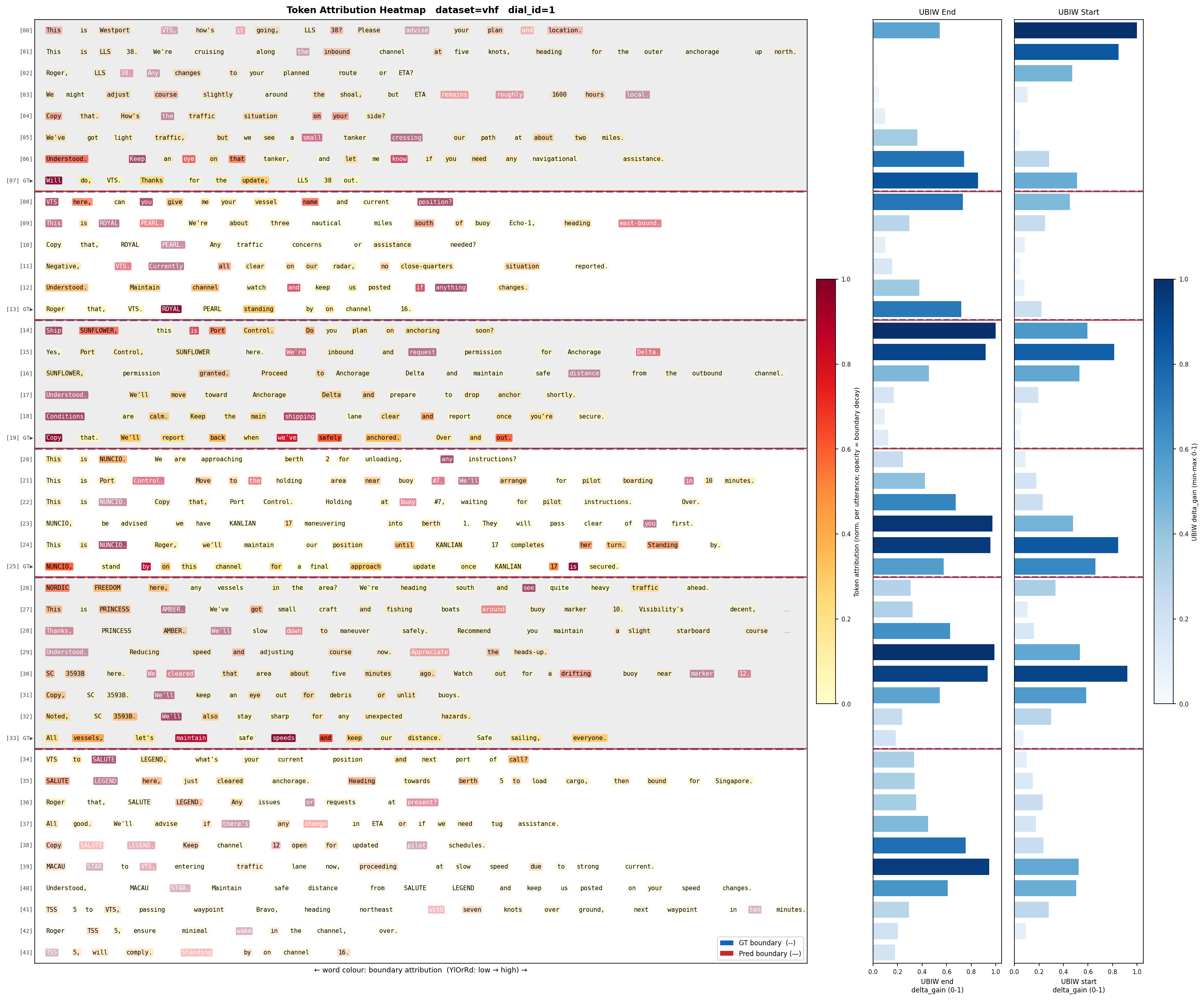}
\caption{Attribution at the token level and informativeness at the utterance level on a VHF example. Lexical cues adjacent to boundaries receive stronger emphasis than less informative interior tokens.}
\label{fig:ubiw}
\end{figure*}

\begin{figure}[!htbp]
\centering
\includegraphics[width=\columnwidth]{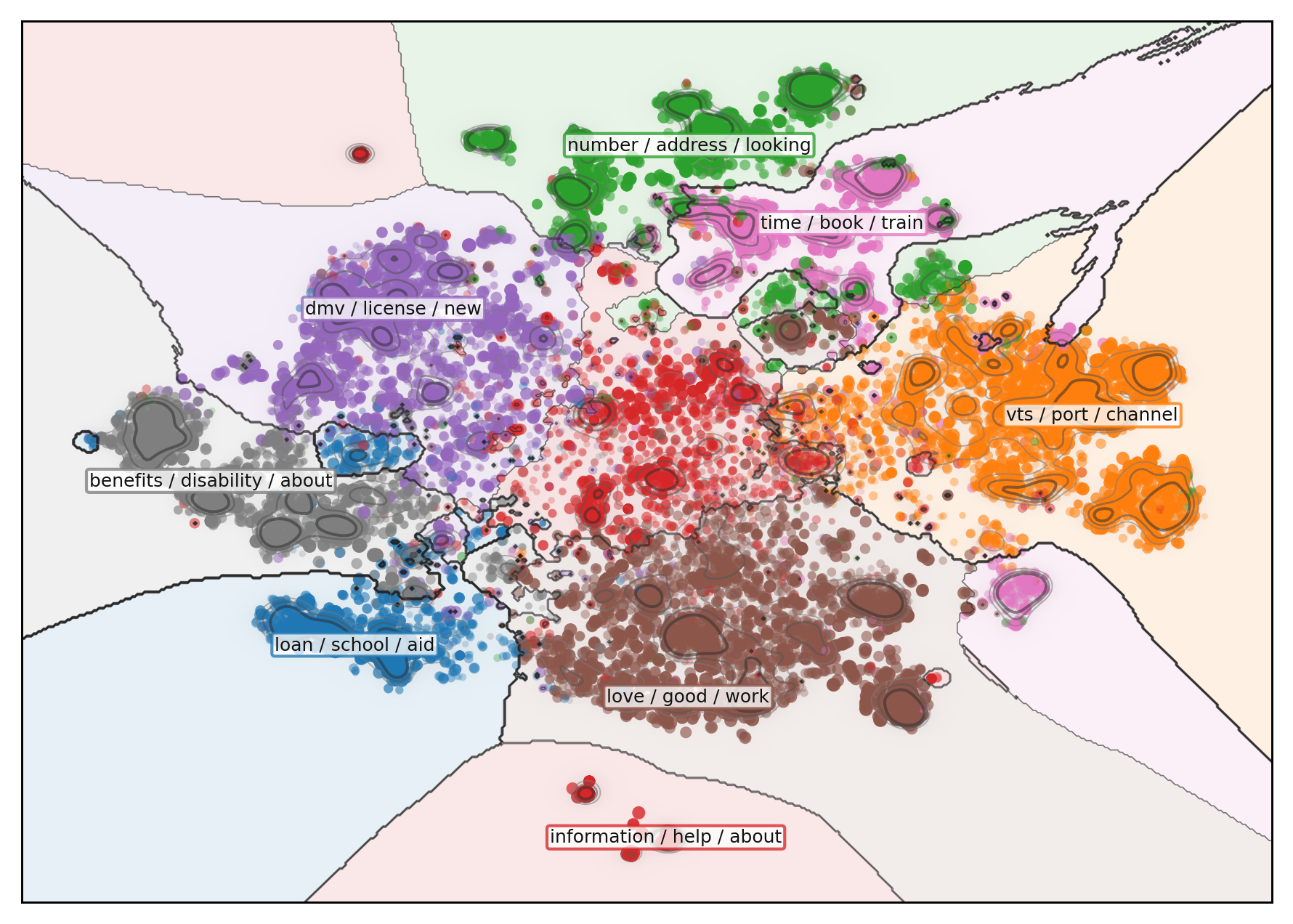}
\caption{Visualization of utterance embeddings with topic-region assignments. The projected topic regions reveal both vocabulary cohesive within topics and lexical triggers sensitive to boundaries. The topic coherence cue is derived from the topic partition, not from the visualization projection.}
\label{fig:kmeans}
\end{figure}

\Cref{fig:ubiw} visualizes boundary attribution at the token level and utterance informativeness for a VHF example. Tokens adjacent to boundaries and high utility procedural phrases receive stronger emphasis than mid-utterance filler content. This matches the intended role of the Lexical Boundary Detector and boundary informativeness weighting. \Cref{fig:kmeans} presents a visualization of utterance embeddings with topic-region assignments. The topic coherence cue keywords are derived from the unsupervised topic partition itself; the projection is used only for visualization. The topic regions, local density patterns, and keywords at the region level indicate that the topic partition captures both vocabulary that is cohesive within topics and lexical triggers sensitive to boundaries, providing the statistical basis for the topic coherence cue.

\section{Unsupervised Pseudo-Segment Induction}
\label{sec:appendix-pseudo}

The pseudo-label pipeline produces training boundaries through three stages. Stage 1 (NSP+TeT initialization): provisional segment boundaries are obtained by running NSP-based coherence scoring and TextTiling peak picking on each training dialogue. The threshold coefficient $\alpha$ is selected on the validation split, then the full training split is cut and 100 training dialogues are sampled. Stage 2 (segment label-summary specification): the LLM receives all provisional segments of a sampled dialogue and outputs one pseudo label and one pseudo summary for each segment. Stage 3 (dialogue reconstruction): target segment lengths are drawn from the empirical training segment-length distribution, and the LLM receives the ordered pseudo label-summary records with these target lengths to generate an entirely new dialogue in one call. This length-controlled reconstruction keeps the synthetic training set close to the original corpus in its structural statistics, so the pseudo data changes the lexical realization while preserving the segment-length distributional profile. Because the segment lengths of the reconstructed dialogue are fixed and validated, the resulting boundaries are known by construction and serve as pseudo-boundary labels. The pipeline does not expose the reconstruction LLM to original dialogue text. LLM configuration and pipeline parameters are listed in \Cref{tab:params}.

The constrained input-output schema of the two LLM-assisted stages is described below rather than reproducing the raw prompts verbatim in the paper body.

\subsection{Segment Label-Summary Specification}

The label-summary stage receives all provisional segments of a dialogue and outputs one pseudo label and one pseudo summary for each segment. The LLM processes the entire provisional dialogue structure at once, preserving segment order and keeping the number of segments unchanged.

\begin{lstlisting}[style=promptstyle, caption={Prompt for segment-level pseudo label-summary induction.}]
[System]
You label each dialogue segment with a pseudo
label and a pseudo summary. Return JSON only.
Do not explain. Return exactly one JSON object
with a pseudo_seg key.
-------------------------------------------------
[User]
Return JSON with exactly one key: pseudo_seg.
pseudo_seg must be an array of objects, one per
dialogue segment. Each object must have exactly two
string keys: pseudo_label and pseudo_summary.
The output must preserve the input segment order.
The output must contain exactly the same number
of segments as the input.
Each output item must correspond to one input
segment and must not merge or split segments.

Input JSON:
{
  "pseudo_seg": [
    {
      "pseudo_seg_index": 0,
      "pseudo_seg_len": 4,
      "utterances": ["...", "...", "...", "..."]
    }
  ]
}
\end{lstlisting}

\subsection{Dialogue Reconstruction from Segment Specifications}

The reconstruction stage generates a complete dialogue from segment metadata alone. The LLM receives only the ordered pseudo label-summary records and target segment lengths sampled from the empirical training distribution, without any original dialogue text. The output contains generated utterances and segment lengths, which are known by construction. This length control is a distribution-matching constraint: each generated segment is forced to have the requested number of utterances so that the reconstructed corpus preserves the original dataset's segment-length characteristics.

\begin{lstlisting}[style=promptstyle, caption={Prompt for dialogue reconstruction from segment metadata.}]
[System]
You generate a full coherent dialogue from
segment specifications. Return JSON only.
Do not explain. Return exactly two keys:
pseudo_utterance and pseudo_seg.
-------------------------------------------------
[User]
Generate one continuous dialogue.
Return JSON with exactly two keys:
pseudo_utterance and pseudo_seg.
pseudo_utterance must be a JSON array of
plain strings.
Do not prefix utterances with speaker
identifiers, ship names, role tags, Speaker A/B,
speaker:, or speaker labels.
pseudo_seg must be a JSON array of integers, and
each integer is a segment LENGTH, not a segment
index.
pseudo_seg must list the utterance counts for
each segment in order.
Each integer must match the requested target
segment length.
The number of segments must be exactly K.
The pseudo_seg array must be exactly
[l_1, ..., l_K].
The total number of utterances must be exactly
sum([l_1, ..., l_K]).
The pseudo_utterance array length must equal the
sum of the pseudo_seg array.

Input JSON:
{
  "pseudo_seg_count": K,
  "target_pseudo_seg": [l_1, ..., l_K],
  "total_target_pseudo_utterance": sum([l_1, ..., l_K]),
  "pseudo_seg_specs": [
    {
      "pseudo_seg_index": 0,
      "pseudo_label": "navigation coordination",
      "pseudo_summary": "two vessels coordinate
                         a safe passing plan",
      "pseudo_seg_len": 4
    }
  ]
}
\end{lstlisting}

The constrained prompts ensure that the LLM produces dialogues with exact segment boundaries. The segment count comes from NSP+TeT initialization, while target segment lengths are sampled from the training split segment-length distribution to keep the reconstructed corpus statistically aligned with the source dataset. The reconstruction LLM sees only topic-level metadata and sampled lengths.

\end{document}